\documentclass[letterpaper, preprint, paper,11pt]{AAS}

\usepackage[inkscapeformat=png]{svg}
\usepackage{mathtools}
\usepackage{gensymb}
\usepackage{float}
\usepackage{graphicx}
\usepackage{xcolor}
\usepackage{tabularx}
\usepackage{adjustbox}

\usepackage{bm}
\usepackage{amsmath}
\usepackage{subfigure}
\usepackage{comment}
\usepackage{overcite}
\usepackage{footnpag}			      	% make footnote symbols restart on each page
\usepackage{amssymb}
\usepackage{subfigure}   
\usepackage{multicol}
\usepackage{multirow}
\usepackage{booktabs}
\usepackage{pstool}             % for changing text in eps graphics
\usepackage[colorlinks]{hyperref}

\usepackage{amsthm}

\PaperNumber{24-378}

\usepackage{soul}
\usepackage[colorinlistoftodos]{todonotes}
\usepackage{pdfcomment}
\begin{document}

\title{Markers Identification for Relative Pose Estimation of an Uncooperative Target }

\author{Batu Candan \thanks{PhD Student, Department of Aerospace Engineering, Iowa State University, IA 50011, USA. email: dukynuke@iastate.edu} 
\ and Simone Servadio\thanks{Assistant Professor, Department of Aerospace Engineering, Iowa State University, IA 50011, USA. email: servadio@iastate.edu}
}

 \maketitle
 
\begin{abstract}
This paper introduces a novel method using chaser spacecraft image processing and Convolutional Neural Networks (CNNs) to detect structural markers on the European Space Agency's (ESA) Environmental Satellite (ENVISAT) for safe de-orbiting. Advanced image pre-processing techniques, including noise addition and blurring, are employed to improve marker detection accuracy and robustness. Initial results show promising potential for autonomous space debris removal, supporting proactive strategies for space sustainability. The effectiveness of our approach suggests that our estimation method could significantly enhance the safety and efficiency of debris removal operations by implementing more robust and autonomous systems in actual space missions. 
% 70/100 words
\end{abstract}
 
\section{Introduction}

The escalating problem of space debris necessitates innovative solutions for identification and removal, particularly for large defunct satellites like ESA's ENVISAT, an inactive Earth observation satellite. In the past ten years, deep learning (DL) has profoundly influenced the development of computer vision algorithms, enhancing their performance and robustness in various applications like image classification, segmentation, and object tracking. This momentum has carried into spacecraft pose estimation, where DL-based methods have begun to surpass traditional feature-engineering techniques as reported in the literature [\citen{pauly, SONG202222, sharmaNeural}], corner and marker detection algorithms such as Shi-Tomasi, Hough Transform  methods [\citen{capuano19, sharmaHough}]. 

CNNs have the edge over feature-based methods primarily due to their enhanced robustness against poor lighting conditions and their streamlined computational demands. However, when it comes to space imagery, the scenario changes due to the distinct challenges such as high contrast, low signal-to-noise ratio, and inferior sensor resolution, which can diminish accuracy. Generally, the scarcity of extensive synthetic space image datasets, crucial for comprehensive CNN training, necessitates the use of networks pre-trained on terrestrial images. To adapt these for space applications, transfer learning is employed, focusing on training only select layers of the CNN [\citen{cassiniRew}]. Detection of the keypoints such as corners involves predicting the 2D projections of specific 3D keypoints from the spacecraft's imaged segments using a deep learning model. These keypoints are usually determined by the spacecraft’s CAD model. In the absence of a CAD model, methods like multiview triangulation or structure from motion are employed to create a 3D wireframe model that includes these keypoints [\citen{chenDL, huo}]. Keypoint location regression technique involves direct estimation of the keypoint positions [\citen{huanDL, hayabusa}]. Segmentation-driven method utilizes a network with dual functions, segmentation and regression, to deduce keypoint locations. The image is sectioned into a grid, isolating the spacecraft within certain grid cells. Keypoint positions are then computed as offsets within these identified cells, enhancing prediction accuracy. Variations of this model have been optimized for space deployment, reducing parameter count without compromising on prediction accuracy [\citen{segment1, segment2}]. Heatmap prediction method represents the likelihood of keypoint locations, from which the highest probability points are extracted as the actual locations. Moreover there is a bounding box approach using the enclosing bounding boxes over the keypoints are predicted along with the confidence scores rather than utilizing the keypoint locations or heatmaps [\citen{heatmap1, piazza}].

On the other hand, estimation involves deducing the value of a desired quantity from indirect, imprecise, and noisy observations. When this quantity is the current state of a dynamic system, the process is termed, where the best estimate is obtained by eliminating noise from the measurements. Moreover, estimation of the relative position and the prediction of the target attitude are crucial for safe proximity operations. This necessitates complex on-board computations at a frequency ensuring accuracy requirements are met. However, the limited computational power of current space processors constrains the estimation processes that can be implemented. Therefore, developing efficient algorithms that minimize computational demands while maintaining necessary performance and reaching the best estimate are crucial for the success of these missions. This estimate is produced by an optimal estimator, a computational algorithm that processes data to maximize a specific performance index, effectively utilizing available data, system knowledge, and disturbance information. For linear and Gaussian scenarios, the posterior distribution remains Gaussian, and the Kalman Filter (KF) is used to compute its mean and covariance matrix. However, practical problems are often nonlinear, leading to non-Gaussian probability density functions. Various techniques address nonlinear estimation problems. One straightforward method is linearizing the dynamics and measurement equations around the current estimate, as done in the Extended Kalman Filter (EKF), which applies KF mechanics to a linearized system. Higher-order Taylor series approximations can extend the EKF's first-order approximation. The Gaussian Second Order Filter (GSOF), for example, truncates the Taylor series at the second order to better handle system nonlinearities. This requires knowledge of the estimation error's central moments up to the fourth order to calculate the Kalman gain. For instance, while the EKF uses first-order truncation requiring covariance matrices, the GSOF requires third and fourth central moments of the state distribution. The GSOF approximates the prior PDF as Gaussian at each iteration and performs a linear update based on a second-order approximation of the posterior estimation error. Other linear filters use different approximations, such as Gaussian quadrature, spherical cubature, ensemble points, central differences, and finite differences [\citen{servadio2021estimation}]. Alternatively, the Unscented Kalman Filter (UKF) employs the unscented transformation to better manage nonlinearities in dynamics and measurements, typically achieving higher accuracy and robustness than the EKF. The UKF uses this transformation for a more precise approximation of the predicted mean and covariance matrix, remaining a linear estimator where the estimate is a linear function of the current measurement. The UKF offers a solution by using the unscented transformation, which avoids linearization by propagating carefully selected sample points through the nonlinear system, providing superior performance in such situations [\citen{servadiojsr, servakoopman}]. 

This study presents a novel approach using image processing from a chaser spacecraft to detect structural markers on the ENVISAT satellite, and a estimation framework utilizing unscented Kalman filtering facilitating its safe de-orbiting. Utilizing high-resolution imagery, the project employs advanced CNN for precise marker detection, essential for the subsequent removal process. The methodology incorporates image pre-processing, including noise addition and blurring, to enhance feature detection accuracy under varying space conditions. Preliminary results demonstrate the system’s efficacy in identifying corner points on the satellite, and ability to keep translational and rotational estimates in appropriate levels promising a significant leap forward in automated space debris removal technologies. This work builds upon recent advancements in space debris monitoring and removal strategies, echoing the urgent call for action highlighted in studies such as [\citen{klinkrad-2006}, \citen{liou-2006}, \citen{servaThreat}], which emphasize the growing threat of space debris and the necessity for effective removal mechanisms [\citen{servadio2024risk}]. Our findings indicate a scalable solution for debris management, aligning with the proactive strategies recommended by the Inter-Agency Space Debris Coordination Committee (IADC) [\citen{mejiakaiser-2020}].

\section{Methodology}

\subsection{Dynamics}
In the following analysis, several assumptions are made to simplify the dynamics modeling of the chaser and target spacecraft. Firstly, it is presumed that the inertia properties of both the chaser and the target are perfectly known beforehand. This assumption simplifies the dynamic modeling by removing uncertainties related to mass distribution and moments of inertia. Secondly, the motion of the chaser spacecraft is considered deterministic. This means that the trajectory and state of the chaser are precisely known and not affected by any form of noise or uncertainties, leading to an idealized model of the chaser's dynamics. Lastly, neither flexible dynamics nor external disturbances are taken into account. The analysis ignores the effects of external factors such as gravitational perturbations, atmospheric drag, or solar radiation pressure. This simplification implies that the translational and rotational dynamics of the spacecraft are decoupled, allowing for independent analysis of these two aspects. These assumptions streamline the analysis by focusing on the primary dynamics without the added complexity of uncertain factors or external influences.

\subsubsection{Absolute Chaser Motion:}
The chaser motion is described by the following equations, where $\mu$ is the gravitational parameter of the Earth, $\Bar{r}$ is the position of the chaser centre of mass with respect to the Earth, and $\theta$ is the true anomaly in the orbit of the chaser.

\begin{equation}
   \ddot{\Bar{r}} = \Bar{r} \dot{\theta}^2 - \frac{\mu}{\Bar{r}^2}
\end{equation}

\begin{equation}
   \ddot{\theta} = -2 \frac{\dot{\Bar{r}} \dot{\theta}}{\Bar{r}}
\end{equation}

\subsubsection{Relative Translational Dynamics:}
The target, ENVISAT, has its relative translational dynamic equations developed with respect to the chaser local-vertical-local-horizontal (LVLH) frame of the chaser. The target relative position, denoted as $\mathbf{r}_r$, and relative velocity, $\mathbf{v}_r$, are defined in the chaser LVLH frame as expressed 

\begin{equation}
   \mathbf{r}_r = x\hat{\mathbf{i}}+y\hat{\mathbf{j}}+z\hat{\mathbf{k}}
\end{equation}

\begin{equation}
   \mathbf{v}_r = \Dot{x}\hat{\mathbf{i}}+\Dot{y}\hat{\mathbf{j}}+\Dot{z}\hat{\mathbf{k}} 
\end{equation}

In this context, \(x\), \(y\), and \(z\) are the three components of the vector \( \mathbf{r}_r \) within the chaser LVLH frame, and \( \hat{i} \), \( \hat{j} \), and \( \hat{k} \) represent the respective unit vectors of the reference frame.  Thus, the equations of motion of the target for its relative translational dynamics are able to be written in the following way.

\begin{equation}
    \ddot{x} = 2\dot{\theta}\dot{y} + \ddot{\theta}y + \dot{\theta}^2x - \frac{\mu(\ddot{r}+x)}{[(\Bar{r}+x)^2+y^2+z^2]^{3/2}} + \frac{\mu}{\Bar{r}^2}
\end{equation}

\begin{equation}
    \ddot{y} = -2\dot{\theta}\dot{x} - \ddot{\theta}x + \dot{\theta}^2y - \frac{\mu y}{[(\Bar{r}+x)^2+y^2+z^2]^{3/2}}
\end{equation}

\begin{equation}
    \ddot{z} = -\frac{\mu z}{[(\Bar{r}+x)^2+y^2+z^2]^{3/2}}
\end{equation}

\begin{figure}[H]
    \makebox[\textwidth][c]{\includegraphics[width=0.9\textwidth]{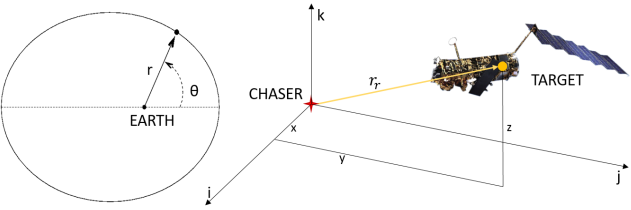}}
    \caption{Visual representation of the elements of translational dynamics}
    \centering
    \label{fig:chasertarget}
\end{figure}

Fig. \ref{fig:chasertarget} represents the visual representation of the translational dynamics.

\subsubsection{Rotational Dynamics:}
The relative orientation of the body-fixed frame on the target with respect to the body-fixed frame of the chaser can be represented with a rotational matrix $\Gamma$. Thus, the relative angular velocity $\omega_r$ in the target body-fixed reference frame is dependent on the angular velocity of the chaser $\omega_c$ and the angular velocity of the target $\omega_t$, both represented in their own body-fixed reference frames.

\begin{equation}
    \omega_r = \omega_t - \Gamma \omega_c
\end{equation}

\begin{equation}
    \dot{\omega}_r = \dot{\omega}_t - \Gamma \dot{\omega}_c + \omega_r \wedge \Gamma \omega_c
\end{equation}

The relative attitude of the target is established by the parametrization of rotation matrix $\Gamma$. The Modified Rodriguez Parameters (MRP) are utilized as the minimal set of three parameters that allows to overcome singularities and to describe every rotation. The quaternion representing the orientation of ENVISAT is expressed as $q$, which consists of four components: $q_0$, $q_1$, $q_2$, and $q_3$. The three-axis angular velocity of ENVISAT is represented by $\omega$, with components $\omega_x$, $\omega_y$, and $\omega_z$.

\begin{equation}
    \mathbf{q} = \begin{bmatrix}
                    \Bar{\mathbf{q}} \\
                    q_4
                \end{bmatrix}
\end{equation}

\begin{equation}
    \Bar{\mathbf{q}} = \begin{bmatrix}
                        q_1 \\
                        q_2 \\
                        q_3
                        \end{bmatrix} = \mathbf{\hat{n}} \sin{\frac{\phi}{2}}
\end{equation}

\begin{equation}
    q_4 = \cos{\frac{\phi}{2}}
\end{equation}

Note that $\mathbf{\hat{n}}$ is the unit vector  corresponding to the axis of rotation and $\theta$ is the rotation angle. MRPs are connected to the quaternions in the following way.

\begin{equation}
    \mathbf{p} = \frac{\Bar{\mathbf{q}}}{(1+q_4)} = \mathbf{\hat{n}} \tan{\frac{\phi}{4}}
\end{equation}

Vector $\mathbf{p}$ is the MRP vector, with dimension of $3\times1$. The kinematic equation of motion are able to be derived by using the target’s relative angular velocity, therefore the time evolution of the MRP is described as following.

\begin{equation}
    \dot{\mathbf{p}} = \frac{1}{4}\left[(1-\mathbf{p}^T\mathbf{p})\mathbf{I}_3 + 2\mathbf{p}\mathbf{p}^T + 2[\mathbf{p}\wedge]\right]\omega_r
\end{equation}

Note that $I_3$ is a $3 \times 3$ identity matrix and $[\mathbf{p}\wedge]$ is a $3 \times 3$ cross product matrix given as following. 

\begin{equation}
    [\mathbf{p}\wedge] = \begin{bmatrix}
                        0 & -p_3 & p_2 \\
                        p_2 & 0 & -p_1 \\
                        -p_2 & p_1 & 0
                        \end{bmatrix}
\end{equation}

The rotation matrix that connects the chaser body-fixed frame and the target body-fixed frame can thus be derived as following.

\begin{equation}
    \alpha_1 = 4 \frac{1-\mathbf{p}^T\mathbf{p}}{(1+\mathbf{p}^T\mathbf{p})^2}
\end{equation}

\begin{equation}
    \alpha_2 = 8 \frac{1}{(1+\mathbf{p}^T\mathbf{p})^2}
\end{equation}

\begin{equation}
    \Gamma(\mathbf{p}) = \mathbf{I}_3 - \alpha_1 [\mathbf{p}\wedge] + \alpha_2 [\mathbf{p}\wedge]^2
\end{equation}

The absolute rotational dynamics of the chaser is described by the torque-free Euler equations. The relative attitude dynamics are obtained by substituting the kinematics relationship in the Euler absolute equations of the target spacecraft.

\begin{equation}
    \mathbf{J}_t \dot{\omega}_r + \omega_r \wedge \mathbf{J}_t \omega_r = \mathbf{M}_{app} -\mathbf{M}_{g} - \mathbf{M}_{ci}
\end{equation}

Note that $\mathbf{J}_t$ is the matrix of inertia of the target, $\mathbf{M}_{app}$ is the apparent torques, $\mathbf{M}_g$ is the gyroscopic torques, and $\mathbf{M}_{ci}$ is the chaser-inertial torques.

\begin{equation}
    \mathbf{M}_{app} = \mathbf{J}_t \omega_r \wedge \Gamma \omega_c
\end{equation}

\begin{equation}
    \mathbf{M}_{g} = \Gamma \omega_c \wedge \mathbf{J}_t \Gamma \omega_c + \omega_r \wedge \mathbf{J}_t \omega_c + \Gamma \omega_c \wedge \mathbf{J}_t \omega_r
\end{equation}

\begin{equation}
    \mathbf{M}_{ci} = \mathbf{J}_t \Gamma \dot{\omega}_c
\end{equation}

\subsection{ENVISAT Satellite Simulation}
\begin{figure}[!ht]
    \makebox[\textwidth][c]{\includegraphics[width=\textwidth]{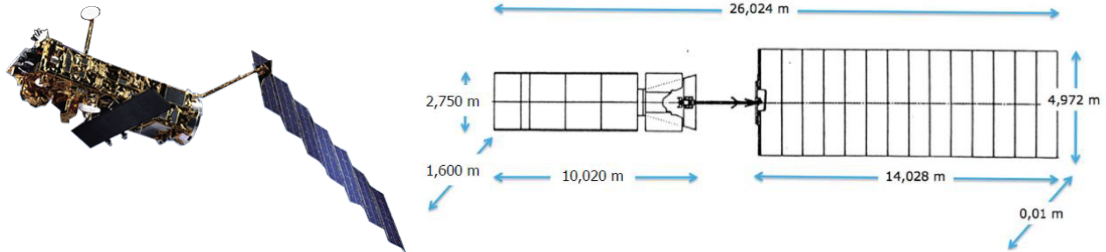}}
    \caption{ENVISAT satellite with its geometrical dimensions}
    \centering
    \label{fig:envisat}
\end{figure}

The simulation of the spacecraft, specifically the ENVISAT satellite, was meticulously conducted in MATLAB. In this controlled environment, we generated detailed models of the satellite's orientation and movement patterns, replicating the complex dynamics encountered in orbit. By leveraging MATLAB's computational tools, we were able to create accurate ground truth data for both the pose and marker locations on the satellite by propagating the equations of motion of the chaser and the target satellites via using Runge-Kutta-Fehlberg method. This simulation process was critical in establishing a reliable dataset that mirrors the real-world conditions the satellite would experience in the space. Fig. \ref{fig:envisat} visualizes the ENVISAT satellite and its geometrical dimensions used within the simulation environment (commas are decimal seperators).

\begin{figure}[!ht]
    \makebox[\textwidth][c]{\includegraphics[width=1\textwidth]{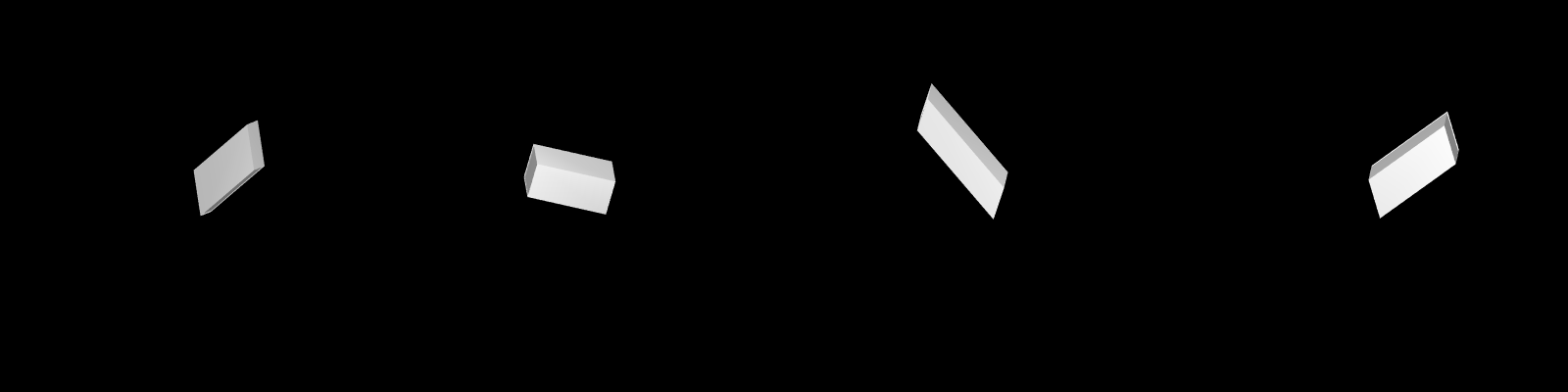}}
    \caption{Simulated images of ENVISAT satellite}
    \centering
    \label{fig:mergedsim}
\end{figure}

Fig. \ref{fig:mergedsim} shows the series of simulated images for the ENVISAT satellite as following. Note that, the solar panel is not included within the simulation but the main parallel-piped body is modelled for assessing the accuracy of the corner detection algorithm and verification. State vector that has 12 components (the relative position between target and chaser centres of mass, relative velocity of the centre of mass, modified Rodriguez parameters for attitude, and angular velocities) and the true measurement of all the corner locations (which would be the ground truth for the corner detection algorithm) are stored after each simulation. 

\subsection{Data Preparation}
Upon completion of the simulation, the generated images, encapsulating the precise pose and marker information of the satellite, were transferred to a Python environment for further processing. The transition from MATLAB to Python was essential for integrating the simulation data with the image processing and machine learning pipeline that follows. 

In Python, one of the initial steps involved converting the RGB images obtained from the simulation into gray-scale. This conversion is pivotal for the subsequent corner detection process, as working with gray-scale images reduces computational complexity and focuses the analysis on the structural information relevant for marker detection. By eliminating the color data, we can enhance the efficiency and accuracy of the CNN in identifying the critical corner locations on the satellite. The streamlined dataset, now optimized for the neural network processing, sets the stage for the effective application of machine learning techniques in detecting the markers essential for the satellite's pose estimation and debris removal strategy.

\section{Convolutional Neural Network}
CNN is a type of deep learning model designed to process data with a grid-like topology, such as images. CNNs use convolutional layers to automatically and adaptively learn spatial hierarchies of features from input data. Key components include convolutional layers, pooling layers, and fully connected layers. The architecture of CNNs has evolved from simpler models like AlexNet to more complex ones like High-Resolution Network (HR.Net). These advancements have significantly enhanced their ability to perform tasks like image and video recognition, object detection, and more. The advantages of using CNNs over traditional neural networks in computer vision include:

\begin{itemize}
\item CNNs weight-sharing mechanism reduces the number of trainable parameters, improving generalization and reducing overfitting.
\item Simultaneous learning of feature extraction and classification layers leads to a well-organized model output that heavily relies on extracted features.
\item Implementing large-scale networks is more straightforward with CNNs compared to other neural network types.
\end{itemize}

The CNN architecture consists of multiple layers, each serving a distinct function:

\begin{itemize}
    \item Convolutional Layer: The core component, using filters (kernels) to process input images and generate feature maps.
    \item Pooling Layer: Sub-samples feature maps to reduce their size while retaining dominant information, using methods like max, min, and global average pooling.
    \item Activation Function: Determines whether a neuron should be activated, mapping inputs to outputs.
    \item Fully Connected Layer: Each neuron connects to all neurons in the previous layer, serving as the classifier.
    \item Loss Functions: Calculate the error between predicted and actual outputs, guiding the learning process.
\end{itemize}

\subsection{CNN Architecture}
\begin{figure}[H]
    \makebox[\textwidth][c]{\includegraphics[width=1\textwidth]{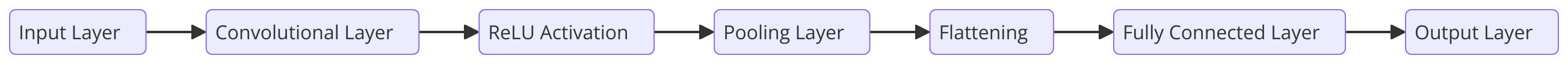}}
    \caption{Overall CNN architectural diagram}
    \centering
    \label{fig:cnndiagram}
\end{figure}

Over the past decade, numerous CNN architectures have been introduced. These architectures have significantly enhanced performance across various applications through structural changes, regularization, and parameter optimization. Notably, major improvements have stemmed from reorganizing processing units and developing new blocks, particularly by increasing network depth. This section examines key CNN architectures from AlexNet in 2012 to the High-Resolution (HR) model in 2020, analyzing features like input size, depth, and robustness to guide researchers in selecting the appropriate architecture for their tasks [\citen{alzubaidi-2021}]. In our work, chosen neural network model is based on L-CNN, a novel neural network designed for comprehensive wire-frame parsing in an end-to-end manner [\citen{lcnn}]. This network comprises four main components: a feature extraction backbone, a junction proposal module, a line verification module, and a connecting line sampling module. Starting with an RGB image as input, L-CNN efficiently outputs a vectorized representation directly, bypassing heuristic methods. The architecture of L-CNN is fully differentiable, allowing for end-to-end training via back-propagation. This capability harnesses the full potential of cutting-edge neural network designs for effective scene parsing. Fig. \ref{fig:cnndiagram} provides a straightforward view to CNN architecture as following.

Input layer is where the raw data (e.g., images) is fed into the network. Convolutional layer applies convolutional filters to the input data to extract features such as edges, textures, and patterns. Each filter creates a feature map by sliding over the input data and performing element-wise multiplications and summations. The rectified linear unit (ReLU) activation block function introduces non-linearity into the network. It replaces all negative pixel values in the feature maps with zero, allowing the network to learn more complex patterns. Pooling layer reduces the spatial dimensions of the feature maps (width and height) while retaining the most important information. Common types of pooling include max pooling, which selects the maximum value from a set of neighboring pixels, and average pooling, which computes the average value. Flattening block converts the 2D feature maps into a 1D vector, which can be fed into the fully connected layers. This step is necessary for transitioning from the convolutional layers to the fully connected layers. Fully connected layer connects every neuron in the previous layer to every neuron in the next layer. It combines the features extracted by the convolutional layers to make final predictions. This layer often includes several fully connected layers for deeper networks. Finally, the output layer produces the final predictions of the network. In classification tasks, this layer typically uses a softmax activation function to output a probability distribution over the possible classes [\citen{ajitReview}].

\section{Unscented Kalman Filter (UKF)}
UKF is an estimation algorithm designed to handle nonlinear systems. Unlike the Extended Kalman Filter (EKF), which relies on linearizing the system and measurement equations using first-order Taylor expansions, the UKF employs the unscented transformation to directly address nonlinearity without linearization. However, it is important to underline that it still employs a linear measurement update [\citen{servadio2020recursive}], which better fits the limited computational power of onboard computers. 

This method involves propagating a set of carefully chosen sample points through the nonlinear system to accurately capture the posterior mean and covariance. The primary advantage of the UKF lies in its ability to provide superior performance in highly nonlinear environments. By avoiding the inaccuracies introduced by linearization, the UKF ensures a more robust and accurate estimation process. In the context of spacecraft pose estimation, particularly for missions involving rendezvous with uncooperative targets, the UKF is invaluable. It improves the prediction of the spacecraft's relative position and attitude, crucial for safe and precise proximity operations. The UKF's effectiveness is demonstrated through its application in various scenarios, such as the European Space Agency's e.deorbit mission, which targets the ENVISAT satellite. The performance of the UKF, evaluated through numerous numerical simulations, highlights its advantages in terms of accuracy, robustness, and computational efficiency, making it a preferred choice for handling the nonlinear dynamics of space missions [\citen{servadio2020nonlinear}]. 

\subsection{Algorithm Steps}
The weights for the sigma points in the Unscented Kalman Filter (UKF) are determined using the parameters \(\alpha\), \(\beta\), and \(\kappa\). These parameters help in controlling the spread and scaling of the sigma points around the mean. The parameter \(\alpha\) determines the spread of the sigma points, \(\kappa\) is a secondary scaling parameter, and \(\beta\) is used to incorporate prior knowledge of the distribution (for Gaussian distributions, \(\beta = 2\) is optimal). The scaling parameter \(\lambda\) is computed as:

\begin{equation}
\lambda = \alpha^2 (L + \kappa) - L
\end{equation}
where \(L\) is the dimensionality of the state vector.

The weights for the mean and covariance of the sigma points are given by:

\begin{equation}
W_m^{(0)} = \frac{\lambda}{L + \lambda}
\end{equation}

\begin{equation}
W_c^{(0)} = \frac{\lambda}{L + \lambda} + (1 - \alpha^2 + \beta)
\end{equation}

\begin{equation}
W_m^{(i)} = W_c^{(i)} = \frac{1}{2(L + \lambda)}, \quad i = 1, \ldots, 2L
\end{equation}

Here \(W_m^{(i)}\) are the weights for the mean. \(W_c^{(i)}\) are the weights for the covariance and \(\lambda\) is the composite scaling parameter. These weights ensure that the sum of the weights is 1, which maintains the consistency of the state estimation process.

\subsubsection{Summary of Parameters}
\begin{itemize}
    \item \(\alpha\): Determines the spread of the sigma points around the mean. Typically a small positive value.
    \item \(\beta\): Incorporates prior knowledge of the distribution. For Gaussian distributions.
    \item \(\kappa\): A secondary scaling parameter, usually set to 0 or 3-L.
\end{itemize}

The UKF algorithm can be summarized in the following steps:

\subsubsection{Initialization}

\begin{equation}
\hat{\mathbf{x}}_0 = \begin{bmatrix}
x_{0,1} \\
x_{0,2} \\
\vdots \\
x_{0,n}
\end{bmatrix}
\end{equation}

\begin{equation}
\mathbf{P}_0 = \begin{bmatrix}
P_{0,11} & P_{0,12} & \cdots & P_{0,1n} \\
P_{0,21} & P_{0,22} & \cdots & P_{0,2n} \\
\vdots & \vdots & \ddots & \vdots \\
P_{0,n1} & P_{0,n2} & \cdots & P_{0,nn}
\end{bmatrix}
\end{equation}

First, the initial state estimate \( \hat{\mathbf{x}}_0 \) and the initial state covariance matrix \( \mathbf{P}_0 \) are defined. These represent the initial guess of the state and its uncertainty.

\subsubsection{Prediction Step}
    \begin{equation}
    \chi_k^{(i)} = 
    \begin{cases} 
      \hat{x}_{k-1} & \text{for } i = 0 \\
      \hat{x}_{k-1} + (\sqrt{(L+\lambda)P_{k-1}})_i & \text{for } i = 1, \ldots, L \\
      \hat{x}_{k-1} - (\sqrt{(L+\lambda)P_{k-1}})_{i-L} & \text{for } i = L+1, \ldots, 2L
    \end{cases}
    \end{equation}
    where \( \chi_k^{(i)} \) are the sigma points, \( \hat{x}_{k-1} \) is the previous state estimate, \( P_{k-1} \) is the previous state covariance, \( L \) is the dimension of the state, and \( \lambda \) is a scaling parameter.

    \begin{equation}
    \chi_{k|k-1}^{(i)} = f(\chi_{k-1}^{(i)}, u_{k-1})
    \end{equation}
    where \( f \) is the state transition function and \( u_{k-1} \) are the control inputs.

    \begin{equation}
    \hat{x}_{k|k-1} = \sum_{i=0}^{2L} W_m^{(i)} \chi_{k|k-1}^{(i)}
    \end{equation}
    where \( W_m^{(i)} \) are the weights for the mean.

    \begin{equation}
    P_{k|k-1} = \sum_{i=0}^{2L} W_c^{(i)} \left( \chi_{k|k-1}^{(i)} - \hat{x}_{k|k-1} \right) \left( \chi_{k|k-1}^{(i)} - \hat{x}_{k|k-1} \right)^T
    \end{equation}
    where \( W_c^{(i)} \) are the weights for the covariance and the process noise covariance is set 0.

\subsubsection{Update Step}
\begin{equation}
\gamma_k^{(i)} = h(\chi_{k|k-1}^{(i)})
\end{equation}
where \( h \) is the measurement function and $\gamma_k^{(i)}$ is the sigma points that are transformed through the measurement function. The predicted measurement mean \(\hat{\mathbf{z}}_k\) is computed as following.

\begin{equation}
\hat{z}_k = \sum_{i=0}^{2L} W_m^{(i)} \gamma_k^{(i)}
\end{equation}

The predicted measurement covariance \(\mathbf{S}_k\) is calculated as:

\begin{equation}
S_k = \sum_{i=0}^{2L} W_c^{(i)} \left( \gamma_k^{(i)} - \hat{z}_k \right) \left( \gamma_k^{(i)} - \hat{z}_k \right)^T + R_k
\end{equation}
where \( R_k \) is the measurement noise covariance. Kalman gain $K_k$ is calculated as following.

\begin{equation}
T_k = \sum_{i=0}^{2L} W_c^{(i)} \left( \chi_{k|k-1}^{(i)} - \hat{x}_{k|k-1} \right) \left( \gamma_k^{(i)} - \hat{z}_k \right)^T
\end{equation}

\begin{equation}
K_k = T_k S_k^{-1}
\end{equation}

Now, the state and the state covariance can be updated as following. 

\begin{equation}
\hat{x}_k = \hat{x}_{k|k-1} + K_k (z_k - \hat{z}_k)
\end{equation}

\begin{equation}
P_k = P_{k|k-1} - K_k S_k K_k^T
\end{equation}

Overall, the UKF stands out as a powerful tool in the realm of nonlinear estimation, offering significant improvements over traditional methods by leveraging its nonlinear transformation capabilities.

\section{Filtering}
The measurement model in filtering combines translational and rotational information. However, the propagation of dynamics can be separated into translational and rotational components, resulting in a faster and more efficient estimation of relative translational states (relative position \( \mathbf{r}_r \) and relative velocities \( \mathbf{v}_r \)) and relative rotational states (MRP, \( \mathbf{p} \) and angular velocities, \( \boldsymbol{\omega}_r \)). This approach allows the state vector, although 12 components long, to be divided into two separate parts of 6 components each, propagating the translational and rotational models in parallel. The filters use a 4th-order Runge-Kutta integrator for propagation. At the start, the required marker positions and chaser absolute states are loaded, and an initial estimate of the relative states, in terms of mean and covariance, is provided. Before beginning the estimation, the filter uses information from the previous step to calculate marker visibility if it is not already given. Depending on the simulation requirements, the filter can operate with the entire set of markers or limit measurements to three markers, with the selection process explained later. Additionally, measurement failures can be included in the simulation. Finally, the estimated relative states are compared with the true states propagated by the dynamics simulator to evaluate filter performance.

\section{Measurement Model}

\subsection{Markers Creation}
Filters require accurate measurements to effectively correct the predicted values and accurately determine the target's attitude. Most filters for space applications depend on camera image processing. In practical scenarios, the image processing software is configured to identify target points in each captured image, known as markers. The software processes the camera image, identifies the marker positions, and then transmits this information to the filter. Selecting markers is a complex task influenced by the target's shape, volume, and color, as the image processing must be rapid. Commonly, target corners are selected as markers, utilizing reliable corner detection algorithms like the Harris-Stephens [\citen{Harris1988ACC}] and Förstner algorithms. Effective interaction between the filter and the image processing software is crucial. After an initial period during which the first measurements are taken and the position error rapidly decreases, the communication should be optimized to expedite marker estimation. Once the filter completes its iterative cycle, it can inform the camera where to search for markers in the subsequent image. This enables the camera software to analyze a smaller image region, reducing the need to process all pixels and focusing only on those near the predicted marker positions. In the context of the ENVISAT relative pose estimation problem, it was decided to use the corners of the main body as markers and track their positions over time. Figure 2 already illustrates the dimensions of the spacecraft. The European Space Agency online sources provide details about ENVISAT’s mass, the location of its center of mass (without propellant), its moments of inertia (without propellant), as well as its geometrical center and volume. Consequently, the marker positions can offer valuable information since the position of each marker is well known with respect to the centre of mass. By tracking the trajectory of these markers, the filter can reconstruct the state of the spacecraft and calculate its relative position and velocity. The main body of ENVISAT, excluding the solar panel, can be modeled as a simple parallelepiped with 8 corners. These corners are selected as filter markers, and their positions relative to the center of mass are known. Each marker is identified by a letter, resulting in markers labeled A, B, C, D, E, F, G, and H.

\subsection{Measurement Equations}
The state vector has 12 components that can be divided in four parts, divided into four equal parts. Each part, composed by three elements, describes one aspect of the attitude on the target satellite, ENVISAT.

\begin{equation}
\mathbf{x} = (x, y, z, \dot{x}, \dot{y}, \dot{z}, p_1, p_2, p_3, w_{r,x}, w_{r,y}, w_{r,z})
\end{equation}
The four key components to be tracked are: the relative position between the centers of mass of the target and chaser, the relative velocity of the center of mass, the Modified Rodrigues Parameters (MRP), and the angular velocities. Therefore, it is necessary to evaluate the marker positions based on the known state to compare the predicted measurements with the actual measurements obtained from the camera system during the update phase of the algorithm. Let \( \mathbf{u} \) represent the position vector of the chaser's center of mass relative to ENVISAT's center of mass. The measurements for each marker are calculated individually as follows: the marker position vector \( \mathbf{v}_i \) is initially expressed in the target's reference frame. This vector is then transformed into the chaser's reference frame by multiplying it with the rotation matrix \( \Gamma \). Finally, the position of each marker relative to the chaser is determined through a simple vector difference.

\begin{equation}
\mathbf{z}_i = \Gamma^T \mathbf{v}_i - \mathbf{u}
\end{equation}

Here, \( \mathbf{z}_i \) represents the position of a marker relative to the chaser's center of mass, and the rotation matrix \( \Gamma \) is derived from the MRP. Note that MRPs are the part of the state vector, so this is the relation between the states and the measurements too.

\subsection{Markers Visibility}
The presented measurement model relies on the positions of the 8 corners of ENVISAT's main body. However, the camera cannot capture all marker positions in a single frame because some markers are obscured by ENVISAT's structure. As a result, the filter cannot use the entire set of markers simultaneously; it must adjust its measurements frame by frame based on the visible markers. Consequently, the size of the measurement vector varies depending on the number of visible markers. Each marker contributes three position components to the observation, so the measurement vector \( \hat{\mathbf{z}} \) will have \( 3 \cdot i \) components, where \( i = 0, \ldots, 8 \).
The requirement for face visibility is determined by the following equation: If the scalar product between the relative position vector of the chaser and the target, and the unit vector perpendicular to the face is negative, it indicates that the face is oriented towards the camera, making the markers on that face visible.

\begin{equation}
\mathbf{u} \cdot \hat{\mathbf{n}}_i < 0 \quad \text{for} \quad i = \alpha, \ldots, \zeta
\end{equation}

The filter uses the state information at the beginning of each observation to predict which markers will be visible in the next step, preparing to receive the correct number of measurements from the camera. It has hardcoded the arrangement of ENVISAT's faces according to the vectors \( \hat{\mathbf{n}}_i \), thereby predicting marker visibility. The filter handles marker visibility in a binary manner, assigning a value of 1 if a marker is visible and 0 if it is not.

\section{Preliminary Results}
\begin{figure}[H]
    \makebox[\textwidth][c]{\includegraphics[width=0.8\textwidth]{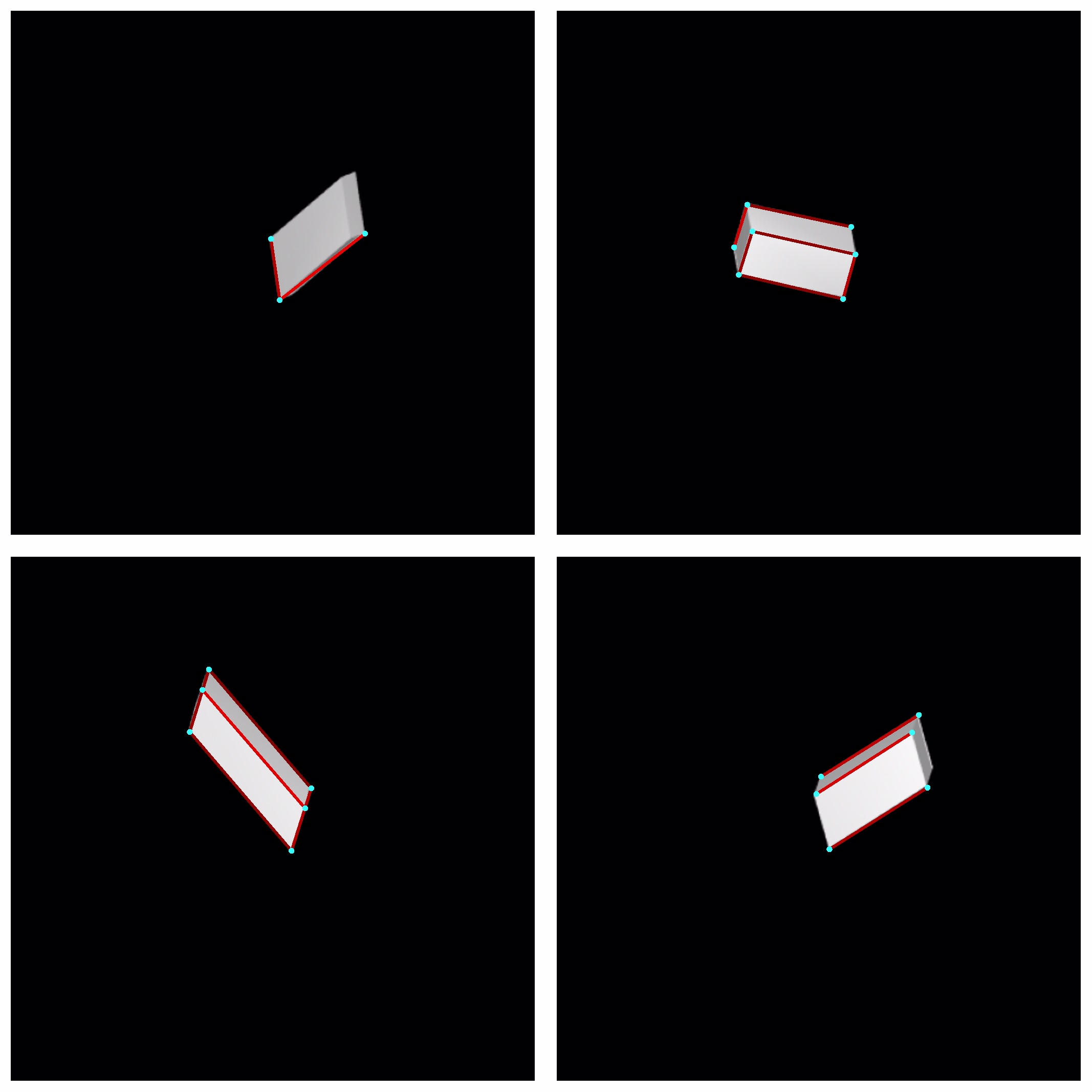}}
    \caption{Preliminary results of the CNN-based corner detection algorithm}
    \centering
    \label{fig:precorner}
\end{figure}

\begin{figure}[H]
    \makebox[\textwidth][c]{\includegraphics[width=0.75\textwidth]{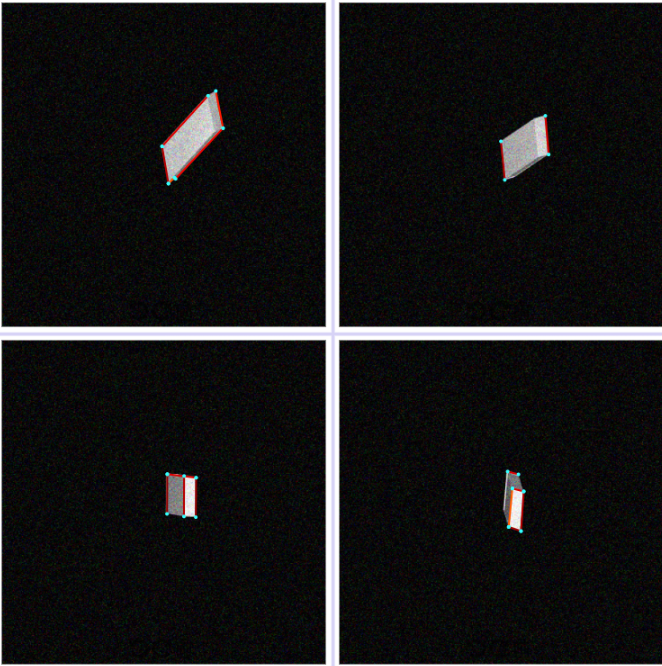}}
    \caption{Preliminary results of the CNN-based corner detection algorithm with noise and blur}
    \centering
    \label{fig:mergedsimnoiseblur}
\end{figure}

Our initial findings demonstrate promising outcomes in the detection and analysis of structural markers on the ENVISAT satellite through processed imagery. This approach not only enhances the robustness of our detection algorithms under varied operational scenarios but also aligns with the typical image quality captured by space-borne sensors. Each image in our dataset underwent a pre-processing phase where Gaussian noise and blur were applied. The addition of Gaussian noise is intended to mimic the electronic noise and sensor imperfections typically found in spacecraft imaging systems. This noise simulates the random variations in pixel intensity that occur due to various factors including thermal effects and the response of the sensor to cosmic radiation. Simultaneously, a Gaussian blur was applied to replicate the slight blurring effect caused by minute focusing discrepancies or motion effects that can occur in a spacecraft’s optical system. Fig. \ref{fig:precorner} visually demonstrates the preliminary results of our algorithm without noise and blur while Fig. \ref{fig:mergedsimnoiseblur} shows the results for the case where the noise and blur are applied. After this step, the offline measurements coming from the CNN algorithm is fed into the UKF framework as the measurements of the filter for the relative pose estimation. It is assumed that a priori knowledge of the geometrical and physical characteristics of both chaser and target is available, rigid body dynamics present, chaser motion is deterministic and there is no external disturbances or control actions. It should be noted that disregarding external disturbances and flexibility allows for the complete separation of the translational and rotational dynamics. Moreover, the proposed measurement model relies on the positions of the 8 corners of ENVISAT's main body. However, the camera cannot detect all marker positions in a single frame because parts of ENVISAT's structure obscure some markers. Consequently, the filter does not operate with the entire set of markers but adjusts its measurements frame by frame based on which markers are visible. 

The visibility and correct association of a corner to its corresponding marker are crucial, as locating more markers tends to enhance the estimation accuracy. Finally, the estimated relative states are compared with the true states propagated by the dynamics simulator to assess the performance of the filters. The findings with Figure 7 indicate a direct correlation between the number of visible markers and the quality of the estimation. Specifically, as the number of observed markers decreases, the accuracy of the position and orientation estimation significantly diminishes. This reduction in markers limits the available data points, leading to higher uncertainty and less reliable estimates. 

\begin{figure}[H]
    \makebox[\textwidth][c]{\includegraphics[width=1.3\textwidth]{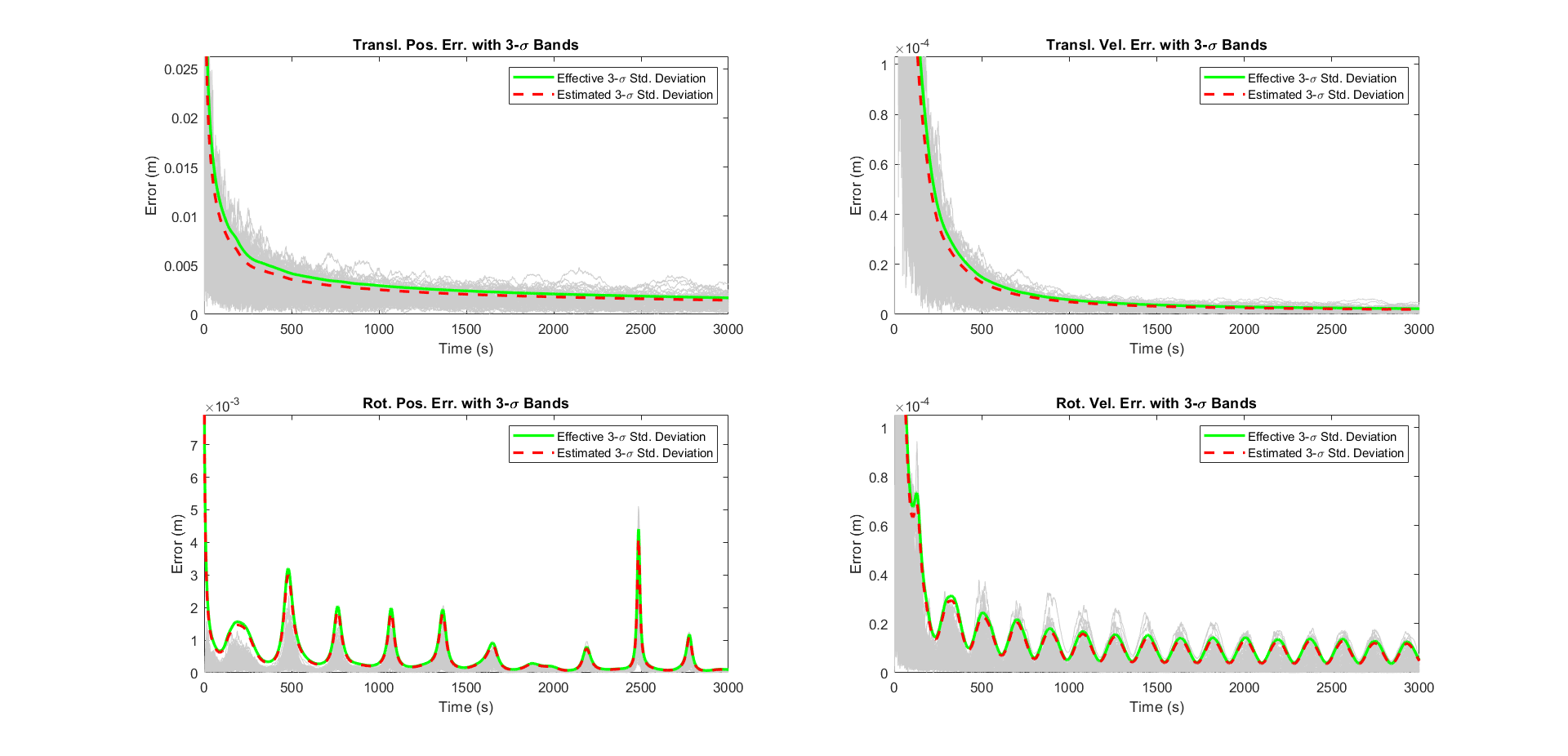}}
    \caption{UKF Monte-Carlo results for $\sigma=0.02$ noise on sensors}
    \centering
    \label{fig:7markers2e-2}
\end{figure}

\begin{figure}[H]
    \makebox[\textwidth][c]{\includegraphics[width=1.3\textwidth]{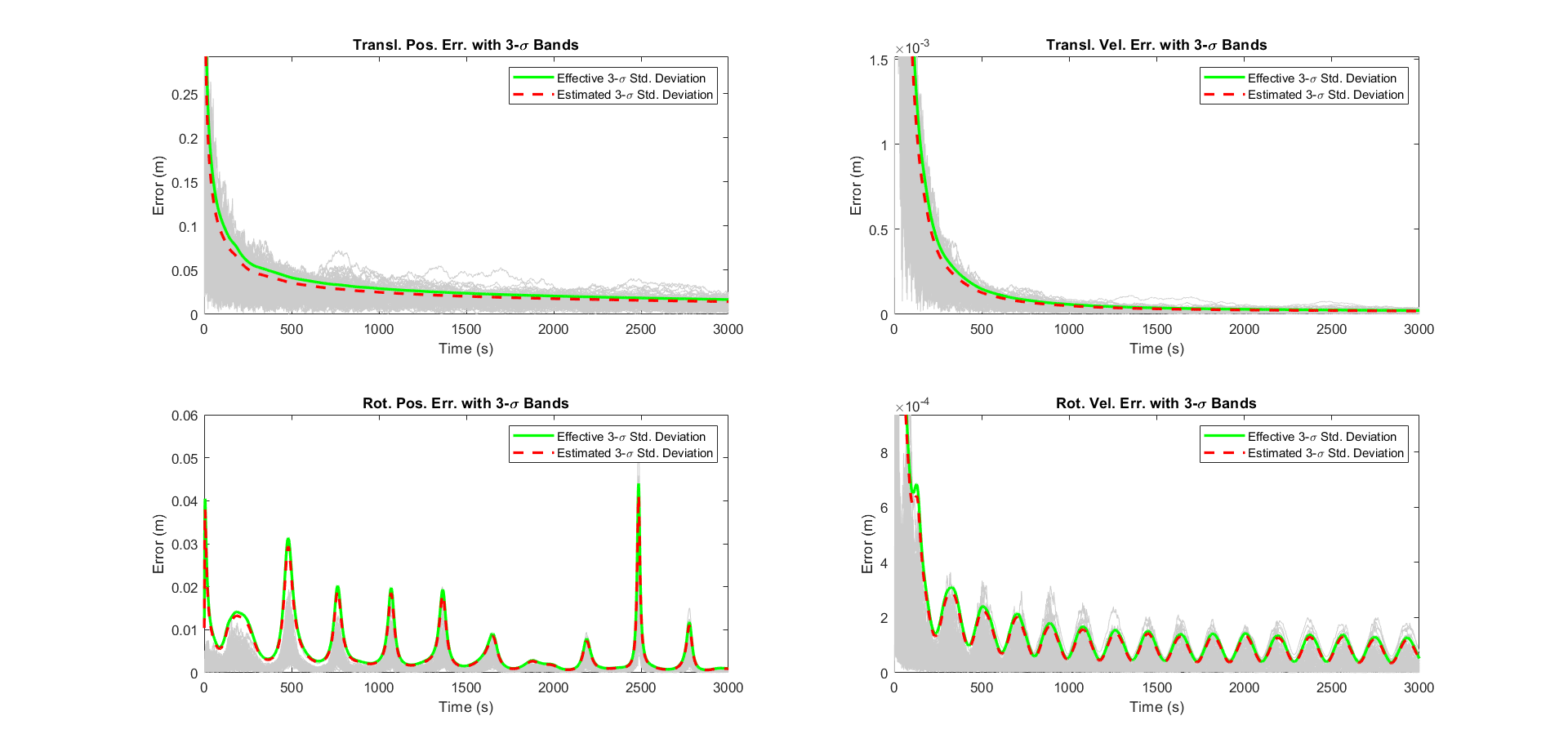}}
    \caption{UKF Monte-Carlo results for $\sigma=0.2$ noise on sensors}
    \centering
    \label{fig:7markers2e-1}
\end{figure}

\begin{figure}[H]
    \makebox[\textwidth][c]{\includegraphics[width=1.3\textwidth]{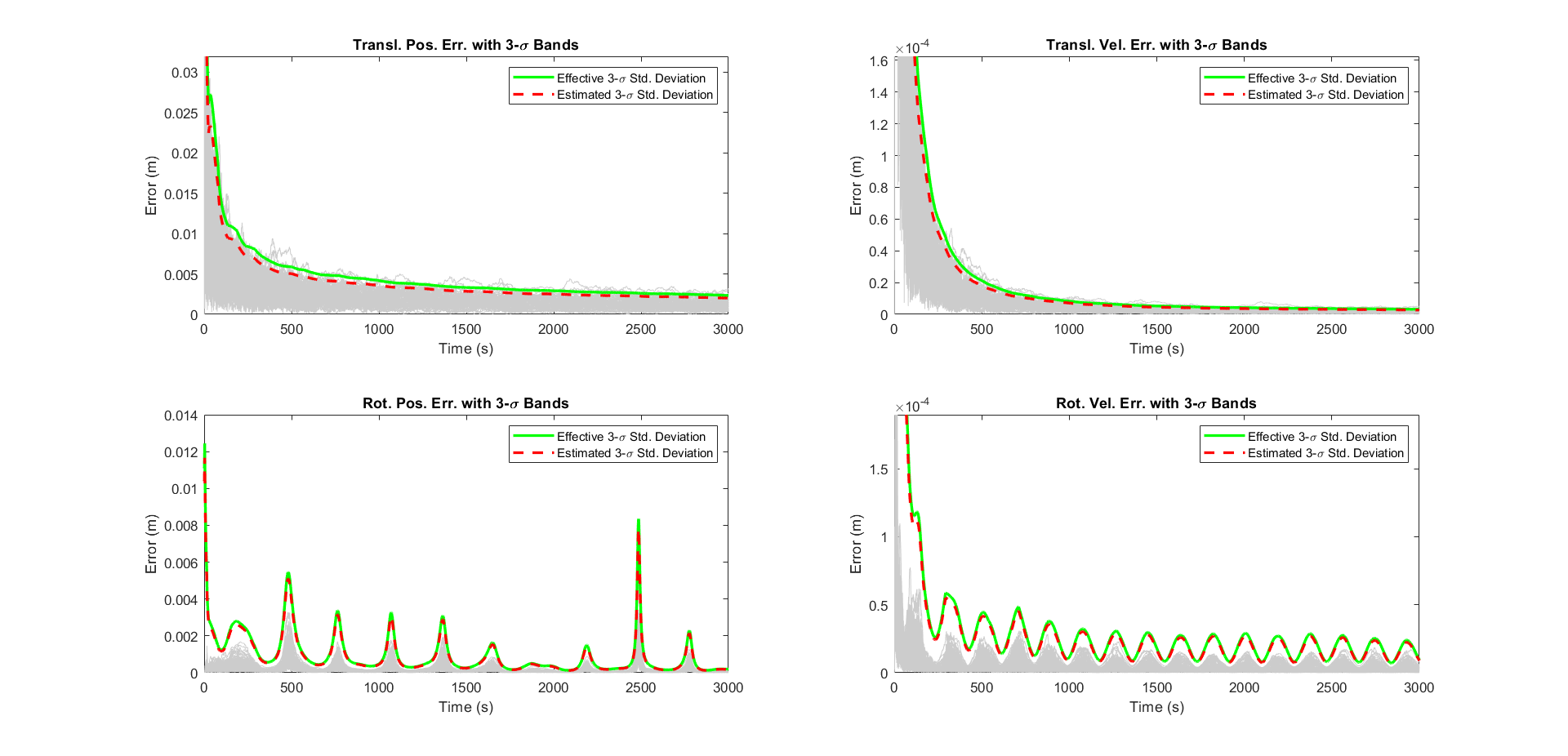}}
    \caption{UKF Monte-Carlo results for $\sigma=0.02$ noise on sensors with only 3 markers visible}
    \centering
    \label{fig:7markers2e-2Noise3Markers.png}
\end{figure}

Moreover, the presence of Gaussian noise and blur on the measurements further exacerbates the degradation in estimation quality. These introduce additional variability and error into the measurements, which in turn affects the overall accuracy of the position and orientation estimates. The difference can be obviously seen on Figure 7 and 8 where the standard deviation of the measurement noise for the sensors is increased from $0.02$ to $0.2$. This dual impact—fewer markers and increased noise—highlights the challenges in accurately determining the satellite's state under sub-optimal conditions. These results underscore the importance of maximizing the number of observed markers and minimizing noise to achieve high-quality estimation of the satellite's relative position and orientation. Finally in Fig. \ref{fig:7markers2e-2Noise3Markers.png}, it is evident that the less markers observed throughout the time period, the estimation accuracy is also decreased. Note that in all Fig. \ref{fig:7markers2e-2}, Fig. \ref{fig:7markers2e-1} and Fig. \ref{fig:7markers2e-2Noise3Markers.png}, the red dashed lines are representing the three sigma values from the error covariance matrix, the green lines are the three sigma values obtained from the sampled errors and the gray lines are basically the Monte Carlo runs during the simulation.

As seen on Fig. \ref{fig:mergedsimnoiseblur}, selected CNN architecture possess the capability to detect corners even in the presence of noise and blur, which are common in real-space environments. This robustness is crucial for applications such as space missions, where the images captured can often be affected by various distortions. The ability of CNNs to accurately identify corners under such conditions enhances their reliability and effectiveness in processing and analyzing visual data from space.

Again from Fig. \ref{fig:7markers2e-2}, Fig. \ref{fig:7markers2e-1} and Fig. \ref{fig:7markers2e-2Noise3Markers.png}, the red dashed lines represent the three-sigma values derived from the estimated error covariance matrix, the green lines show the effective three-sigma values obtained from the sampled errors are close to each other and encapsulating the Monte Carlo errors. Consequently, the UKF has performed well, maintaining estimation accuracy despite the challenges posed by reduced marker observations and other uncertainties in the simulation such as the measurement noise on the sensors. The error spikes which are visible around some points especially for the rotational position and velocity errors. The fundamental reason behind this phenomenon is the positioning of the satellite on the camera frame. However, it is important to state that the filter robustness and accuracy are highly sensitive to the filter frequency, which is set as 1 Hz for this study. The filter tends to diverge below this frequency threshold while it stays convergent above this threshold due to the sampling rate and discretization errors.

\section{Conclusions}
In conclusion, our work successfully demonstrates the capability of a modified CNN to detect corner locations on the ENVISAT satellite, marking a significant advancement in the field of space debris monitoring and removal. Through the use of advanced image processing techniques and neural network architectures, we have established a robust method for identifying crucial structural markers on satellite imagery, which are vital for the planning and execution of debris removal missions. Looking ahead, the detected corner location data present a valuable asset for enhancing navigational accuracy and operational efficiency in space debris management. In future work, we plan to integrate the CNN module which gives corner location data within an UKF framework online to further refine the pose estimation and tracking of space objects. This integration aims to overcome the limitations of current tracking systems by providing more accurate and reliable state estimation under the complex and dynamic conditions of space environments. Moreover, the UKF is going to be improved by trying to implement online adaptation of the measurement noise matrix rather than setting it into a constant matrix. The potential for this advanced approach to significantly improve the precision of space debris tracking and removal operations is immense, contributing to safer space exploration and sustainability efforts in the orbital domain. Future work also should focus on incorporating stochastic absolute values into the estimation process and enhancing the interaction between the camera system and the filter. Introducing stochastic elements can account for uncertainties and variabilities in absolute measurements, thereby improving the robustness and accuracy of the state estimation. Additionally, optimizing the interaction between the camera system and the filter can lead to more efficient processing, allowing the filter to guide the camera in focusing on specific regions of interest, thereby reducing computational load and enhancing real-time performance.

% \section*{Acknowledgments}

% The authors want to acknowledge the support of this work by the Air Force’s Office of Scientific Research under Contract Number FA9550-22-1-0092. 

\bibliographystyle{ieeetr}
\bibliography{references}

\end{document}